\documentclass{sig-alternate}

\usepackage{epsfig}
\usepackage{graphicx}
\usepackage{amsmath}
\usepackage{amssymb}
\usepackage{algorithm}% http://ctan.org/pkg/algorithm
\usepackage{algpseudocode}% http://ctan.org/pkg/algorithmicx
\usepackage{url}
\usepackage[tight,footnotesize]{subfigure}
\usepackage{multirow}
\usepackage{array}
\usepackage{color}
\usepackage[normalem]{ulem}
\usepackage{eurosym}
\usepackage[hidelinks]{hyperref}

\DeclareUrlCommand\ULurl{%
  \renewcommand\UrlLeft{\uline\bgroup}%
  \renewcommand\UrlRight{\egroup}}

\DeclareUrlCommand{\bulurl}{}

\usepackage{xcolor}
\hypersetup{
    colorlinks,
    linkcolor={red!50!black},
    citecolor={blue!50!black},
    urlcolor={blue!80!black}
}

\begin{document}

\CopyrightYear{2017} 
\setcopyright{rightsretained} 
\conferenceinfo{MM '17}{October 23-27, 2017, Mountain View, California, USA} 
%\isbn{978-1-4503-3603-1/16/10}
%\doi{http://dx.doi.org/10.1145/2964284.2973807}

\numberofauthors{2}

\author{
	\alignauthor \makebox[.7\linewidth]{Tam V. Nguyen}\\
	\affaddr{Department of Computer Science\\ University of Dayton}\\
	\email{tamnguyen@udayton.edu}
		\alignauthor \makebox[.7\linewidth]{Luoqi Liu}\\
	\affaddr{Department of ECE\\ National University of Singapore}\\
	\email{liuluoqi@u.nus.edu.sg}
}

%
% --- Author Metadata here ---
\conferenceinfo{ACM MM}{'17 Mountain View, CA, USA}
\CopyrightYear{2017} % Allows default copyright year (20XX) to be over-ridden - IF NEED BE.
%\crdata{0-12345-67-8/90/01}  % Allows default copyright data (0-89791-88-6/97/05) to be over-ridden - IF NEED BE.
% --- End of Author Metadata ---

\title{Smart Mirror: Intelligent Makeup Recommendation and Synthesis}

\maketitle
\begin{abstract}
The female facial image beautification usually requires professional editing softwares, which are relatively difficult for common users. In this demo, we introduce a practical system for automatic and personalized facial makeup recommendation and synthesis. First, a model describing the relations among facial features, facial attributes and makeup attributes is learned as the makeup recommendation model for suggesting the most suitable makeup attributes. Then the recommended makeup attributes are seamlessly synthesized onto the input facial image.  
\end{abstract}



\section{Introduction}

Nowadays, facial makeup has been indispensable for most modern women. As stated in L'Oreal annual report \cite{loreal}, the worldwide cosmetics market has reached \euro $205$ billion in $2016$ along with $4\%$ annual increase. Different from other commercial products, makeup product is person-dependent. The lack of personalized recommendation and effect visualization functionality hinders the cosmetic products' growing for online shopping business. 

How to imitate the physical makeup process and synthesize the effects of these products has attracted the interest of research society \cite{MM12,TOMM13,guo2009digital,scherbaum2011computer,tong2007example} in the past years. Most of the studies utilize image pairs with before-and-after makeup effects. Tong \textit{et al.} \cite{tong2007example} extracted makeup from before-and-after training image pairs, and transferred the makeup effect defined as ratios to a new testing image. Meanwhile, Kristina Scherbaum \textit{et al}. \cite{scherbaum2011computer} used 3D image pairs of before-and-after makeup, and modelled makeup as the ratio of appearance. Guo and Sim \cite{guo2009digital} considered makeup effect existing in two layers of the three-layer facial decomposition result, and makeup effect of a reference image is transferred to the target image. Recently, Li et al.~\cite{Jianshu} presented a deep learning model to capture the underlying relations between facial shape and attractiveness. These works focus on the facial geometric feature while ignoring skin texture information.

\begin{figure}[!t]
	\begin{center}
		\includegraphics[width=1.0\linewidth]{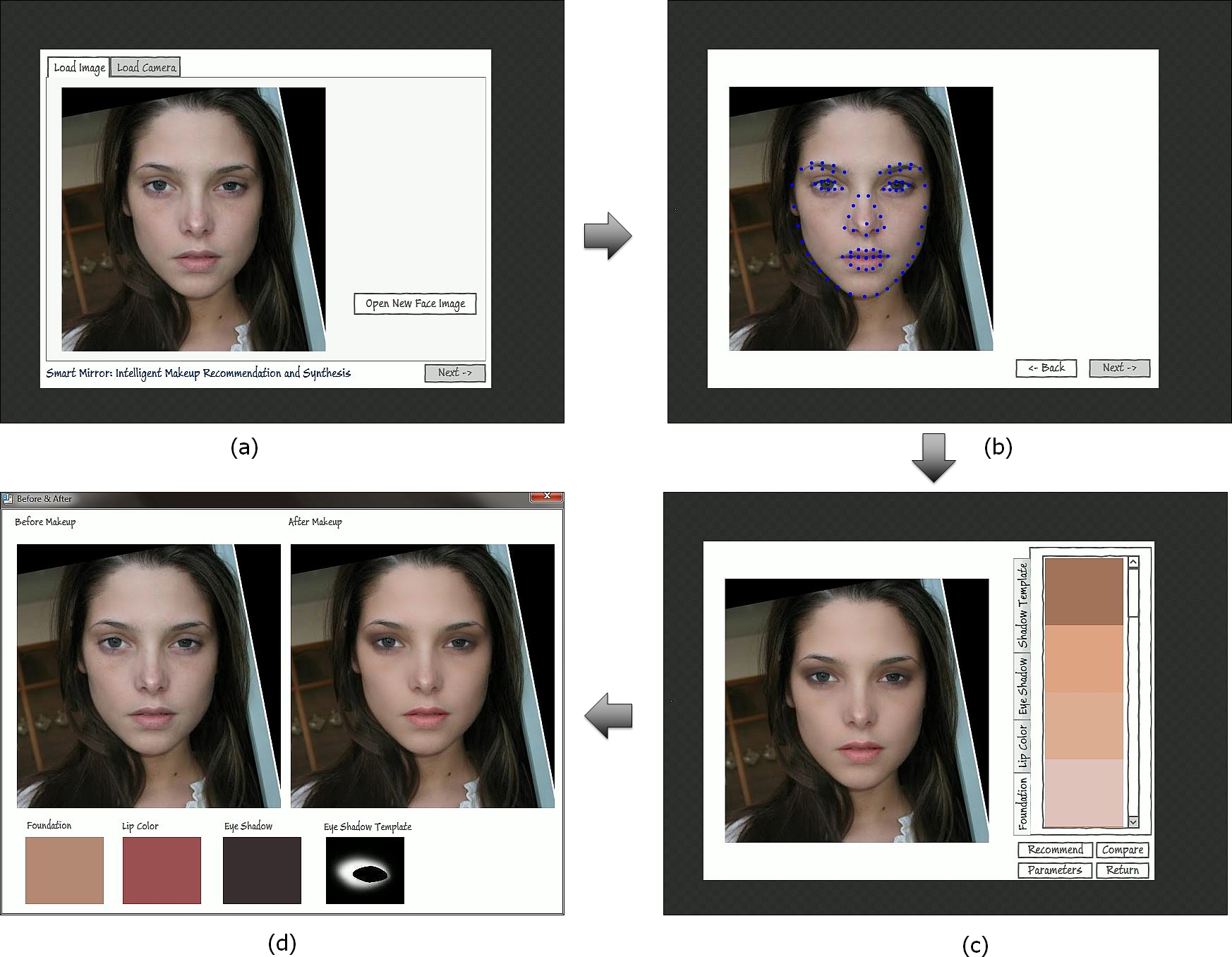}
	\end{center}
	\vspace{-3mm}
	\caption{The user interface of our system: (a) the test facial image, (b) facial feature extraction, i.e., landmark point extraction, (c) makeup recommendation and synthesis, (d) comparison of the test facial image before and after makeup synthesis.}
	\label{fig:gui}
	\vspace{-4mm}
\end{figure}

The aforementioned works are however still limited. First, they do not provide personalized makeup effect/products recommendation function, and the synthesized makeup is unnecessary to be suitable. Second, most of these methods adopt the global transformation approach; however, a good facial makeup effect can be generally decomposed into several individual effects like eye shadow, skin color and lip color. Therefore, in this work, we develop a practical system which can personalize the makeup product recommendation along with a function of visualizing the effect on an input facial image without makeup. The  two functionalities, namely \emph{recommendation} and \emph{synthesis}, can supplement each other, and can significantly facilitate both in-store or online shopping experience. Our user-friendly interface is shown in Figure~\ref{fig:gui}. Additionally, a short video 
about our system is available at \url{https://www.youtube.com/watch?v=tpY-FZtb-Cs}.

% The system can facilitate the following scenarios: 1) a lady without much experience on makeup may feel confused with various types of products, and based on the analysis of her facial attributes from her face photo, our system can help sellers to recommend most suitable products to this lady customer, and 2) for a customer who already selected some products, our system can imitate the physical makeup process to help her foresee the makeup effect. 

%Therefore, in this work, we propose a novel system, which is user-friendly, for automatic facial makeup recommendation and synthesis.

%It should be noted that there exist some commercial softwares to help women simulate the makeup effects, such as Photoshop and Taaz \footnote{http://www.taaz.com/}. But for most of them, heavily manual labeling work is required.

\section{Proposed System}
\subsection{Beauty Makeup Dataset Collection}
Makeup products make great commercial benefits among woman customers, however there exists no public dataset for academic research. Most previous works \cite{tong2007example,scherbaum2011computer,guo2009digital} are example-based or only work on a few selected samples. Chen and Zhang \cite{chen2010benchmark} released a benchmark for facial beauty study, but their focus is geometric facial beauty, not facial makeup. 

Thus, we first collected an image dataset, Beauty Makeup Dataset (BMD), from both professional makeup websites and popular image sharing websites (\textit{e.g.} taaz.com and flickr.com) using key words such as \emph{make up}, \emph{cosmetics}, and \emph{celebrity makeup}. The initially downloaded $\sim500,000$ images are very noisy and contain many non-face images. Thus, we use a commercial face detector \footnote{\label{fn:repeat}OMRON, OKAO vision.\\  http://www.omron.com/r\_d/coretech/vision/okao.html} to detect face and locate $87$ facial key points in the images. Only images with faces in frontal pose, high resolution and with high face and landmark detection confidences are retained. The remaining $\sim10,000$ images are carefully checked to remove faces without obvious makeup effects or with non-uniform illuminations. The final $500$ images are in good quality and with obvious makeup effects.  Then all faces are aligned and decomposed into different regions and the makeup effect analysis is performed for each region. More specifically, we focus on 1) the analysis of eye shadow and exploit spectral matting based methods to extract eye shadow templates; and 2) the modeling of colors of foundation, eye shadow and lip via color clustering method.

% Additionally, another $100$ images without makeup are also collected for testing purpose. Finally all the training and testing images are cropped and aligned with two eye positions as shown in Figure \ref{fig:db}. 

%We first collect a makeup database of $500$ images from celebrities and models with various makeup styles from professional makeup websites and image sharing websites (\textit{e.g.} taaz.com and flickr.com).

%\begin{figure}[!t]
%	\begin{center}
%		\includegraphics[width=1.0\linewidth]{img/bmd.png}
%	\end{center}
%	\vspace{-4mm}
%	\caption{Some exemplar images from the BMD database. Images in the left four columns are from training makeup set, while images in the right column are from non-makeup set for testing purpose.}
%	\label{fig:db}
%	%\vspace{-4mm}
%\end{figure}

\begin{figure}[t]
	\begin{center}
	\includegraphics[width=0.9\linewidth]{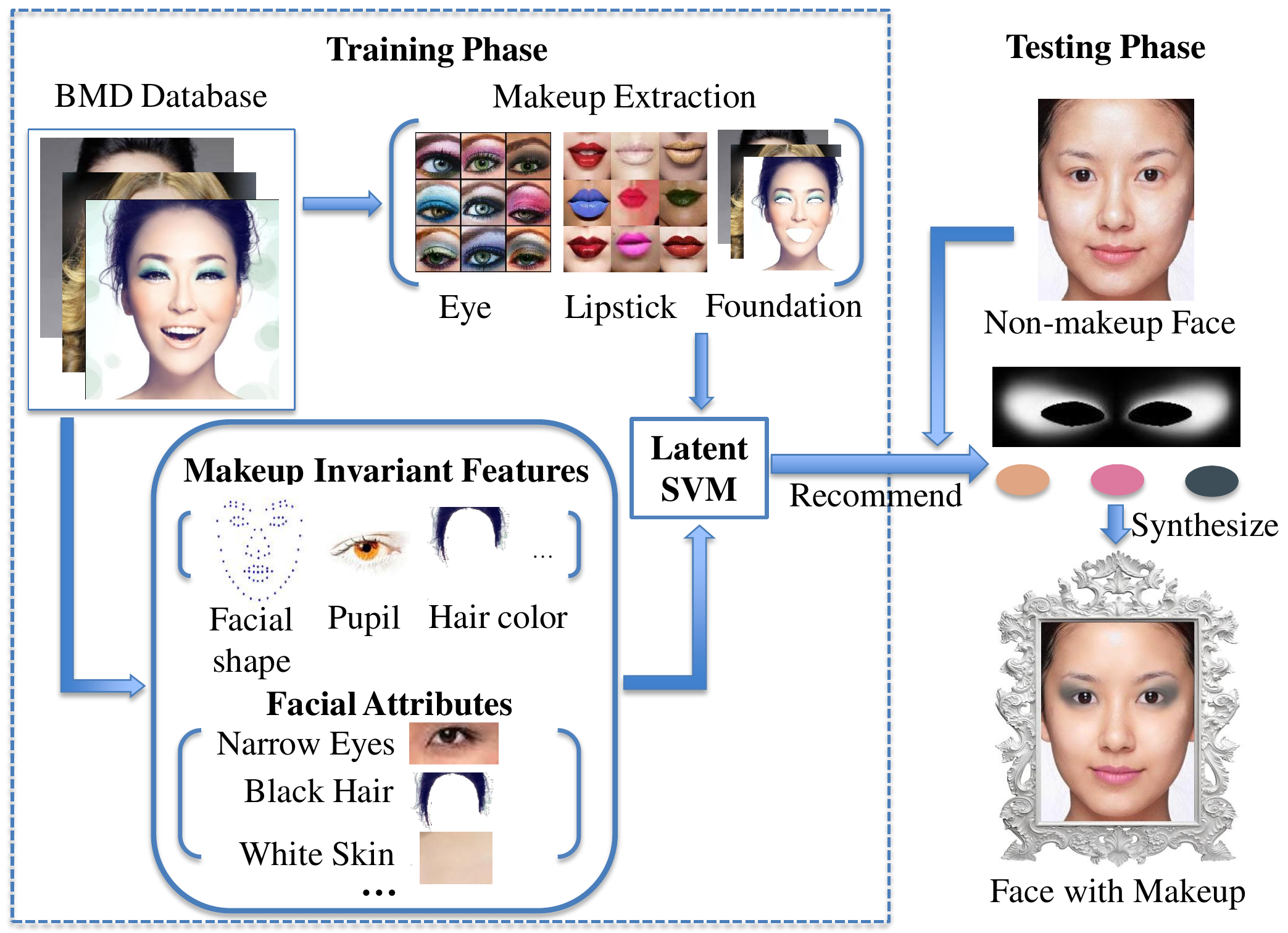}
	\end{center}
	\vspace{-5mm}
	\caption{The flowchart of our proposed makeup recommendation and synthesis system. 
	}
	\label{fig:right_top}
	%\vspace{-3mm}
\end{figure}

\subsection{Makeup Recommendation and Synthesis}
\begin{figure}[t]
	\begin{center}
		\includegraphics[width=\linewidth]{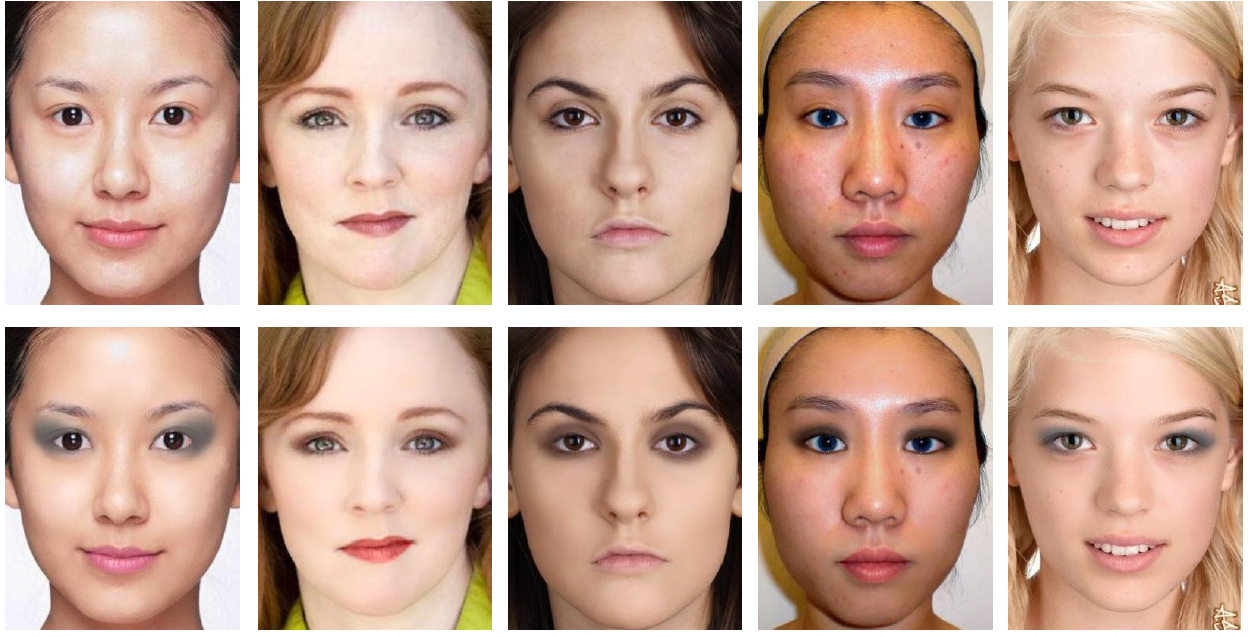}
	\end{center}
		\vspace{-6mm}
	\caption{The results of our makeup synthesis: before-makeup images are listed in the first row, and after-makeup ones are in the second row. Best viewed in $\times 3$ size of original color PDF file.}
	\label{fig:moreresults}
\end{figure}

In this work, we propose a novel system for automatic facial makeup recommendation and synthesis. The flowchart of our system is illustrated in Figure \ref{fig:right_top}. For facial makeup recommendation, makeup-invariant facial features and attributes are designed to describe the intrinsic traits of faces, and a latent SVM model \cite{wang2010discriminative} is learned from our BMD dataset to explore the relation among makeup-invariant facial features, attributes and makeup effects for personalized makeup recommendation. In the synthesis mode, the system recommended makeup products are seamlessly applied onto the testing face for visualizing effects. In particular, we process different facial parts (skin, eyes and lip) separately. Edge preserved filtering~\cite{he2010guided} is applied to imitate the physical effect of foundation, and then different colors are merged into respective regions. The makeup synthesis results are shown in Figure \ref{fig:moreresults}.  A short video about our system is available at \url{https://www.youtube.com/watch?v=tpY-FZtb-Cs}.

% 

%\section{Demonstration Setting}
%For the demonstration, we plan to introduce our system in both makeup recommendation and synthesis. Within this demo, the interested participants are encouraged to try the system with their own captured photos as well as try various additional effects.

\section{Conclusion}
In this work, we showcase a practical system of automatic personalized makeup recommendation and synthesis. It performs the beautification task in a learning-based manner and generates natural makeup results. We hope our work would attract more interesting research works in this area. In the future work, we are planning to focus on more detailed facial regions, \textit{i.e.}, eyebrow and eyelash. In addition, we shall further extend the proposed framework to handle makeup recommendation and synthesis in videos.

\small
\bibliographystyle{abbrv}
\bibliography{egbib} %% That's all folks!
\end{document}